\documentclass[onecolumn, nofootinbib, aps, pra, 11pt]{revtex4-1}

\usepackage{hyperref}
\usepackage[pdftex]{graphicx}
\usepackage{subfig}
\usepackage{amsmath}

\begin{document}

\newcommand{\be}{\begin{equation}}
\newcommand{\ee}{\end{equation}}

\title{Symmetry constrained machine learning}
\date{\today}



\author{Doron L. Bergman}
\affiliation{MegaWatt Technologies LLC, San Jose, CA}

\begin{abstract}%
Symmetry, a central concept in understanding the laws of nature, has been used for centuries in physics, mathematics, and chemistry, to help make mathematical models tractable. Yet, despite its power, symmetry has not been used extensively in machine learning, until rather recently. In this article we show a general way to incorporate symmetries into machine learning models. We demonstrate this with a detailed analysis on a rather simple real world machine learning system - a neural network for classifying handwritten digits, lacking bias terms for every neuron. We demonstrate that ignoring symmetries can have dire over-fitting consequences, and that incorporating symmetry into the model reduces over-fitting, while at the same time reducing complexity, ultimately requiring less training data, and taking less time and resources to train.
\end{abstract}

\maketitle

\section{Introduction}
\label{Sec:Intro}


Symmetry, together with understanding the relevant microscopic degrees of freedom and interactions, are the core ingredients behind our understanding of many body systems in the physical world.


In statistics and computer science, traditionally there has been far less use for symmetry than in the physical sciences,
since the problems these fields were solving often did not posses any symmetrical structures. The machine learning community has begun to explore the uses of symmetry, but there are many more scenarios where it could prove a practical advantage. The recent emergence of convolutional neural networks applied to computer vision~\citep{lecun1989generalization, lecun1999object, lecun1990handwritten, lecun1998gradient, krizhevsky2012imagenet, lee2009convolutional, lecun2015deep}, where translational symmetry of images is used to constrain the form of neural network models, is a first demonstration of the vast potential in building symmetries into machine learning models.


One of the prime challenges for any sufficiently rich model is over-fitting. The physics approach to over-fitting has been to build a more constrained theory, with as few free parameters as possible. The statistics community has not had that luxury, and so they have attacked the over-fitting problem with various methods. Symmetry provides a fruitful avenue to constrain rich models, eliminating superfluous free parameters. By lowering the number of free parameters, models are not only less prone to over-fitting, they also require less data to train, and can be trained and run predictions faster.


There are many obvious symmetries in vision problems - translation, rotations, and mirror images - essentially most portions of affine transformations~\citep{gens2014deep, dieleman2016exploiting, cohen2016group, henriques2016warped}. Even some forms of approximate symmetry exist~\citep{kiddonsymmetry} - the fact that sentences in some languages are still somewhat understandable even if the words are re-arranged, suggests natural languages have an approximate word permutation invariance. This may explain why bag of word methods are often successful, despite not retaining the full information about the word order\footnote{Bag of words methods represent text by the set of words or phrases used in it.}. 

Given the benefits of combining symmetry into machine learning models, it would be useful to have a simple way to incorporate symmetries into any machine learning model. In this paper we address this problem. In short, if inputs are transformed into features that are invariant under the full symmetry group, then the model outputs will automatically be invariant as well. Are we missing any information in this way? No. Any feature that is not invariant under symmetries of the system, will spoil the model invariance, at least for some set of weights. A successful symmetry invariant feature map will be a truncated set of features that suffices to capture enough of the information to make the model successful.

In order to demonstrate the importance and power of symmetry invariant feature maps\footnote{A feature map is a function which maps an input data vector to a vector space to be consumed directly by a machine learning model. For instance, starting with inputs $x_{1,2}$, a feature map could generate the product $x_1 x_2$.}, we consider in this paper a very simple example - recognizing hand written digits using neural networks with bias terms absent. In Sec.~\ref{Sec:model}, we describe the toy model we work with to demonstrate our ideas. In Sec.~\ref{Sec:empirics} we describe results of our empirical computer experiments confirming the mathematical theory, and demonstrating the power and ease of implementing symmetry invariant feature maps. In Sec.~\ref{Sec:general_theory} we show how fitting neural network models without symmetry invariance enforced, leads to shallow directions in the fitting parameter space. Finally we provide some closing remarks in Sec.~\ref{Sec:conclusions}.




\section{Toy model}\label{Sec:model}
\subsection{Neural network model}\label{Sec:dumb_NN}

In this section we describe the neural network model used for demonstrating the utility of symmetry invariant feature maps. The neural network model will be used on the UCI ML hand-written digits dataset\footnote{The original repository for this dataset~\citep{Dua:2017} is \url{http://archive.ics.uci.edu/ml/datasets/Optical+Recognition+of+Handwritten+Digits}} available with the scikit-learn python machine learning library~\citep{scikit-learn}.

For the sake of simplicity, we will assume that our images are comprised of pixels with grayscale values scaled to the range $[-1,+1]$. We shall also use a hyperbolic tangent function as the nonlinear neuron model, rather than a logistic function, thus mapping the internal features of a neural network (NN) to the range of values $[-1, +1]$ as well. The neurons in this case are odd functions $\tanh(-x) = - \tanh(x)$. Finally, we shall not use any biases in any neuron in the system. Denoting the outputs from layer $n$ by $z^{(n)}_i$, the NN is built out of neurons performing the following function
\be
z^{(n+1)}_i = g\left(\sum_j w^{(n)}_{i, j} z^{(n)}_j\right)
\; ,
\ee
where $g(z)$ in our case is $\tanh(z)$. Note the absence of bias terms here - with bias terms the neuron action would be 
$ z^{(n+1)}_i = g\left(b^{(n)}_i + \sum_j w^{(n)}_{i, j} z^{(n)}_j\right) $. The first input layer takes in the image pixels $z^0_i = x_i$. For classifying the digits, a final softmax layer produces 10 outputs which are the normalized probabilities for the image to be any one of the 10 digits
\be\label{softmax_1}
p_{\alpha} \equiv P(y=\alpha | x) \sim \exp\left[ \sum_i u_{\alpha, i} z^{(N-1)}_i \right] = {\tilde p}_{\alpha}
\ee
with the normalization
\be\label{softmax_2}
p_{\alpha} = \frac{{\tilde p}_{\alpha}}{\sum_{\alpha} {\tilde p}_{\alpha} }
\; .
\ee

It must be noted at this point that ignoring the softmax layer for the moment, a neural network with one hidden layer of arbitrary size is a universal function approximator~\citep{hornik1989multilayer}. Our restriction to the squashing function $\tanh$, together with eliminating the biases, reduces the function space the NN can approximate to functions odd in the inputs - this is no longer a universal function approximator.

\subsection{Issues with symmetry}

The NN model above, with just $N=3$ layers, can achieve rather high classification accuracy of the 10 digits (see Sec.~\ref{Sec:empirics} below).
It is trained on UCI ML hand-written digits dataset~\citep{Dua:2017} available with the scikit-learn python machine learning library~\citep{scikit-learn} of roughly black digits on a white background.

We note in passing that with the exception of the input layer to the neural network, at each level we can approximate a bias term, even in this architecture. In the input neuron layer, we take the sum of all the input grayscale values $\sum_i x_i$, and multiply it by a very large weight $w_0 \rightarrow -\infty$. As long as the the images are majority white $\sum_i x_i < 0$, and this neuron will always output $z_0^{(1))} = g\left( w_0 \sum_i x_i \right)\approx 1$. If the images are majority black, we can achieve the same effect with $w_0 \rightarrow +\infty$. Once there is a bias term available as an input to the second layer, another neuron in the 2nd layer can similarly approximate a bias term for the 3rd neuron layer, and so on throughout each subsequent layer $z_0^{(n+1)} = g\left( w_0 z_0^{(n)} \right)\approx 1$. This fact is probably why the neural network architecture, even without biases, is sufficiently general to train effectively on this dataset.

Next, let us challenge the system. Consider the images with the grayscale inverted, such that the digits are now roughly white on 
a black background. The inverted images still represent the same digits - the grayscale inversion transformation is therefore a symmetry of the problem. The geometric contrast is exactly the same, and so we would hope that a good machine learning system captures the geometric features of the digit themselves, and would be able to handle the inverted colors case well.

One can test this empirically (see Sec.~\ref{Sec:empirics} below), and find that the classification accuracy is quite poor, in some cases going below the accuracy of random selection (!). We explore the source of this poor performance, and prove that it is a general effect, rather than a mere anecdote.

Under the grayscale inversion symmetry, $x_i \rightarrow - x_i$, every layer except the final softmax layer (Eqs.~\eqref{softmax_1},\eqref{softmax_2}) is odd in its 
input. Therefore under this symmetry we get all the neuron values flipping their sign $z^{(n)}_i \rightarrow - z^{(n)}_i$. Finally, the unnormalized probabilities transform as
\be
{\tilde p}_{\alpha} \rightarrow \exp\left[ - \sum_i u_{\alpha, i} z^{(N-1)}_i \right] = 1/{\tilde p}_{\alpha}
\; .
\ee
The un-normalized probabilities are inverted. For any given image in the original dataset, the digit the model thinks has the highest probability to match the image, is also the digit with the lowest probability to match the inverted image. For every image the model identifies correctly, the inverted image gets identified incorrectly.

Let us divide the original sample set into two subsets - one for which the model correctly identifies the digit $A$, and $B$ for the remaining samples for which the model identifies the wrong digit. The accuracy rate of the model over this dataset is 
\be
R = \frac{|A|}{|A| + |B|}
\; .
\ee
All the images in $A$, when inverted, will yield the wrong digit from the model. Denoting the subset of the images for which the model identifies the digit correctly on the inverted image as ${\bar A}$, and the remaining as ${\bar B}$, we note that $A \subseteq {\bar B}$. The accuracy of the model on the inverted images is then
\be
{\bar R} = \frac{|{\bar A}|}{|{\bar A}| + |{\bar B}|} = 1 - \frac{|{\bar B}|}{|{\bar A}| + |{\bar B}|} \leq 1 - R
\; .
\ee
Written more symmetrically as $R + {\bar R} \leq 1$ this demonstrates that the accuracy over the original dataset directly constrains the accuracy that can be had on the inverted images. If the original model is accurate more than half the time over this dataset, $R>0.5$, then ${\bar R} \leq 1 - R < 0.5$. Higher accuracy on the original dataset comes directly at the expense of accuracy on the inverted images.

A model such as the one we are considering here would usually be trained on the original dataset, with only black digits on a white background. Once confronted with the reality of poor performance on the inverted images, a common approach is to simply add the additional manipulated samples to the training dataset. This effectively means that one trains not on images, but on orbits of the corresponding symmetry group.

However given the demonstrated $R + {\bar R} \leq 1$ constraint, if the training fully converges onto the best model, then that would suggest 
the constraint is saturated at $R + {\bar R} = 1$, and if the images and the inverted images are given equal weight in determining the model, one would expect $R = {\bar R} = 0.5$, a rather limited ceiling on the potential accuracy of the model.

This is not a very happy state of affairs. 

The weakness of this model stems from two sources - the lack of biases for each neuron, and the features we choose to work with. If we were to re-introduce the biases, what we have proved above no longer holds strictly, but since the accuracy is a smooth function of the biases, if all the biases are infinitesimally small, and if the NN is rather shallow (in practical applications this is nearly always the case) the upper limit on the accuracy can be only infinitesimally larger than in the vanishing biases case $R + {\bar R} \leq 1 + O(\epsilon)$. If the NN is deep this argument will not hold. If one is to understand a deep NN as performing many renormalization group iterations (RG)~\citep{mehta2014exact}, then the NN should be able to take an infinitesimal deviation and amplify it to a sizable deviation. A deep enough NN would be capable of using infinitesimally small biases to violate $R + {\bar R} \leq 1$ significantly, rather than infinitesimally.

In the next section, we consider an alternative approach, where rather than re-introducing the biases, we change the features we feed to the model. In particular, we choose features that are invariant under the inversion symmetry.

\subsection{Incorporating symmetry}\label{Sec:symmetry_invariant_features}

What if we could construct the system to begin with to have the inversion symmetry built in to the model?
The simplest way to do this is to map the inputs (the pixels of the image) $x_i$ onto features that are invariant under the inversion symmetry.
Let us consider a generic smooth feature mapping. It can be expanded in a Taylor series
\be
f({\bf x}) = \sum_{n_1, n_2, \ldots} A_{n_1, n_2, \ldots} x_1^{n_1} x_2^{n_2} \ldots
\; .
\ee

The feature maps that are invariant under the inversion symmetry are those functions even in the inputs $x_i$ - only the even power terms are allowed
\be
f({\bf x}) = A^{(0)} +  \sum_{i, j} A^{(2)}_{i, j} x_i x_j + \sum_{i, j, k, \ell} A^{(4)}_{i, j, k, \ell} x_i x_j x_k x_\ell + \ldots
\; .
\ee
The lowest order feature maps that are non-trivial are the quadratic maps
\be
f({\bf x}) = A^{(0)} +  \sum_{i, j} A^{(2)}_{i, j} x_i x_j
\; .
\ee

If instead of the $x_i$ inputs, we fed the inversion symmetry invariant features $\chi_{i,j} = x_i x_j$ into the NN model, we would enforce the classification to be perfectly invariant under the inversion symmetry.

However, given $x_{1 \ldots M}$ pixels, the number of $\chi_{i,j}$ features is $M(M+1)/2$, a far larger number of features than the $M$ features $x_i$. Using the same neural network architecture with a much larger number of input features is not a fair comparison - there would be many more free parameters using these features than in the original model. Generally a model with more free parameters can better fit a given dataset, albeit with an increased danger of over-fitting.
To level the playing field we will choose a subset of the possible features with a similar number of free parameters. 

A rather natural guess is to choose $x_i^2$ as the symmetry invariant features. However, this choice suffers from a significant part of the information encoded in the inputs $x_i^2$ being lost - the signs of the inputs. In the extreme case where all pixels are either white or black, $x_i = \pm 1$, and $\forall i x_i^2 = 1$, leaving the features $x_i^2$ devoid of any information at all. The signs by themselves are not invariant under the inversion symmetry, but the relative sign between different pixels are
\begin{equation}
\frac{sign(x_i)}{sign(x_j)} = \frac{x_i x_j}{\sqrt{x_i^2 x_j ^2}}
\; .
\end{equation}
It is this information that we shall strive to include in the symmetry invariant features.

A simple choice that will definitely capture all the relative signs, as well as at least some information about the magnitudes of the input values is
\be
x_i x_{i + {\hat e}_1}
\; ,
\ee
where $i + {\hat e}_1$ denotes the pixel immediately to the right of the pixel $i$. For the edge pixels, we wrap the the index arithmetic around the boundaries of the image, essentially acting as if it were wrapped around a cylinder. The products $x_i x_{i + {\hat e}_1}$ contain information similar to a gradient of the image pixels since when both pixels are pure black or pure white, the product is +1, and when they are opposite black and white colors, it is -1. If the image were made of only extreme grayscale values - only black and white, an image made up of the $\chi_i$ features, would show the boundaries between the black and white regions of the original image.

Using the gradient-like features when comparing performance with a NN using the original inputs as features, may be un-fair, since gradient-like features inform the machine learning system about the local nature of the problem. Instead, we consider yet another set of features. We pick a random permutation of the inputs $P$, and use it to construct precisely $M$ features
\begin{equation}\label{random_features}
\chi_i = x_i x_{P(i)}
\; .
\end{equation}
These features do not inform the machine learning model of the local nature of the problem, are invariant under the inversion symmetry, and retain relative sign information whenever $i \neq P(i)$.

We proceed to test the different approaches in a numerical experiment.

\begin{table}
\begin{tabular}{l l l l | l}
	biases & features & training set & test set & accuracy \\ 
	\hline
	No bias & $x_i$ & $X_{train}$ & $X_{test}$ & 0.84 \\
	 &  &  & $-X_{test}$ & 0.001 \\ 
	\cline{3-5}
	 &  & $\pm X_{train}$ & $X_{test}$ & 0.12 \\
	 &  &  & $-X_{test}$ & 0.09 \\ 
	\hline
	\hline
	With bias & $x_i$ & $X_{train}$ & $X_{test}$ & 0.81 \\
	 &  &  & $-X_{test}$ & 0.02 \\ 
	\cline{3-5}
	 &  & $\pm X_{train}$ & $X_{test}$ & 0.68 \\
	 &  &  & $-X_{test}$ & 0.69 \\ 
	\hline
\end{tabular}
\caption{Accuracy measurements of the different models, training and testing sets, with the original inputs as features. Symmetry invariance is not enforced on any of the models.}
\label{table:results_table_no_symmetry}
\end{table}

\begin{table}
	\begin{tabular}{l l l l | l}
		biases & features & training set & test set & accuracy \\ 
		\hline
		No bias & $x_i^2$ & $X_{train}$ & $X_{test}$ & 0.65 \\
		\cline{2-5}
		 & $x_i x_{i + {\hat e}_1}$ & $X_{train}$ & $X_{test}$ & 0.84 \\
		\cline{2-5}
		 & $x_i x_{P(i)}$ & $X_{train}$ & $X_{test}$ & 0.81 \\ 
		\hline
		\hline
		With bias & $x_i^2$ & $X_{train}$ & $X_{test}$ & 0.66 \\
		\cline{2-5}
		 & $x_i x_{i + {\hat e}_1}$ & $X_{train}$ & $X_{test}$ & 0.87 \\	
		\cline{2-5}
		 & $x_i x_{P(i)}$ & $X_{train}$ & $X_{test}$ & 0.82 \\	
		\hline
	\end{tabular}
	\caption{Accuracy measurements of the different models, training and testing sets, with symmetry-invariant features. Symmetry invariance guarantees that the exact same results are obtained on the inverted test sets $-X_{test}$.}
	\label{table:results_table_symmetric}
\end{table}

\section{Numerical experiment}\label{Sec:empirics}

The dataset we use~\citep{Dua:2017} consists of $8 \times 8$ images, and is enlarged by a factor of 5 by duplicating and shifting all the images 1 pixel to the right, left, up, and down, in addition to the original images. This procedure is applied to the dataset before it is fed to any one of the models we compare here.

We compare the performance of a variety of NN models. They all share the same architecture - a fully connected 3-layer neural network model with 10 hidden neurons in the 1st layer, 5 in the second layer, and a softmax layer as the 3rd. 

\begin{figure}
\centering
\includegraphics[width=1.0\linewidth]{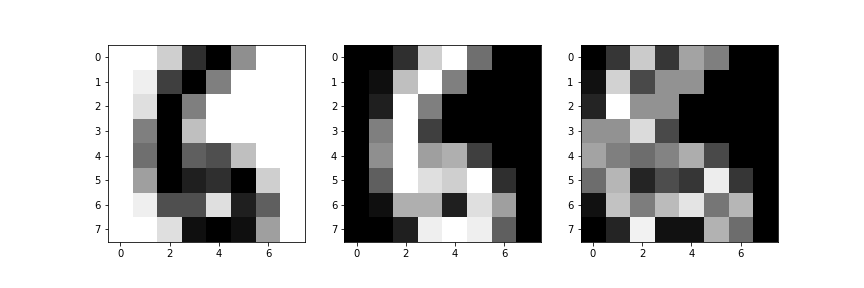}
\caption{A sample image (handwritten figure 6) from the dataset (left), the image after the inversion transformation (middle), and the gradient-like $x_i x_{i + {\hat e}_1}$ feature image symmetric under the inversion transformation (right).}
\label{fig:six_cubed}
\end{figure}

We use neural networks both with and without bias terms - naturally those with bias terms have a few more free parameters. The NN models are used with a variety of different features. The original inputs $x_i$ are used as non-symmetry invariant features, and in this case we train the NN models on a number of different versions of the data. We train either on the original dataset (denoted by $X_{train}$), or a combination of the original dataset, together with the inverted copy of that dataset $-X_{train}$. For all other cases we use features that are symmetry invariant - $x_i^2$, $x_i x_{i + {\hat e}_1}$, and $x_i x_{P(i)}$. Symmetry invariant models we need train only on $X_{train}$, the feature map acting as an additional layer before the NN model receives the data. In Fig.~\ref{fig:six_cubed} we depict one sample, and how it evolves under the different feature transformations we use.

We calculate the accuracy on independent test datasets denoted by $X_{test}$, as well as on its inverted copy denoted by $-X_{test}$. The results for the NN models without symmetry invariant features are enumerated in Table~\ref{table:results_table_no_symmetry}. The results for the NN models using the various symmetry invariant features we propose are enumerated in Table~\ref{table:results_table_symmetric}.

A number of findings are noteworthy. First, the $R + \bar{R} \leq 1$ limit we proved in the previous section clearly holds in our numerical experiment with the models using non-symmetry invariant features in Table~\ref{table:results_table_no_symmetry}. Second, even a neural network model with the biases included generalizes extremely poorly to the inverted samples, when trained only on the original dataset. The same model with biases included does somewhat better when trained on both the original dataset, and the inverted copy of it, but still the accuracy is rather disappointing, and far lower than what the same model can achieve when trained and tested only on the original dataset of black digits on a white background. The NN models using the symmetry invariant features fare best of all.

Our findings suggest that using symmetry invariant features may at the very least result in similar generalization quality to the standard practice of training on a transformed copy of the dataset, and may surpass it to a significant degree.

\section{Universal theory}\label{Sec:general_theory}

\subsection{Fitting parameter degeneracy}\label{Goldstone_modes}

As we have mentioned above, the standard practice of dealing with symmetry is to create symmetry-transformed copies of samples, and add them to the training set. This approach is very appealing because of how straightforward it is. In our toy model, we found that this approach gave a less accurate result than using symmetry invariant features. In this subsection, we shall explore how this standard practice effects the fitting process.

We mention in passing, that if one does not sample transformed versions of the training set, and the model is not invariant to the symmetries, the model will have no information about the symmetry, and will most likely over-fit such that transformed versions of the same examples the model was trained on will yield low predictive accuracy. this certainly is what we have found in the toy model we employed in the previous sections.

Fitting parameters involves minimizing an appropriate utility function
\begin{equation}
\Omega({\bf \theta}) = \sum_i L(y_i, {\hat y}_i)
\end{equation}
where $L(y, {\hat y})$ is the an appropriate cost (or loss) function, such as $(y - {\hat y})^2$, and ${\hat y}_i$ are the predictions of the model
\begin{equation}
{\hat y}_i = f({\bf x}_i; {\bf \theta})
\; .
\end{equation}
Here ${\bf \theta}$ are the model parameters, and $i$ denote the different data samples. In addition to a loss function there may or may not be a regularization term added to the utility function, we shall ignore that for now.

The standard practice, taken to its logical conclusion, amounts to applying each member of the symmetry group on all the samples, and adding them to the training dataset. The utility function then becomes
\begin{equation}\label{standard_practice_fitting}
\Omega({\bf \theta}) = \sum_i \sum_{g \in G} L(y_i, f(U(g) {\bf x}_i; {\bf \theta}))
\; ,
\end{equation}
where $g \in G$ denotes a sum over all the elements in the symmetry group $G$, and $U(g) {\bf x}_i$ denotes the transformation of the vector ${\bf x}_i$ under the group element $g$.

The sum $\sum_{g \in G}$ renders the utility function itself invariant under any symmetry transformations ${\bf x} \rightarrow U(g) {\bf x}$. As a result, in general the minimum of the utility function will not be unique, but rather will span an entire orbit of the symmetry group, and will most likely be in the largest orbits - those of the same number of elements as the symmetry group itself. If the symmetry group is a continuous one, then the minima will form a continuous manifold. This may cause significant numerical convergence issues - the flat directions in parameter space may confuse a stopping criteria, and cause a gradient based algorithm to continue roaming parameter space rather than stop. Similar phenomena in the modeling of MRI data have been recently uncovered by \citet{novikov2018rotationally}. These flat directions, are akin to Goldstone modes appearing in the context of continuous symmetry breaking~\citep{nambu1960quasi, goldstone1961field, goldstone1962broken}.

It is useful to see how the degeneracy in fitting parameter space arises explicitly, especially to understand how things change when we use symmetry invariant features.

Consider a model such as a NN model. The first layer of a NN consists of an affine transformation of the input features. We can write such models in the form
\begin{equation}
{\hat y} = \Psi( {\hat w} \cdot {\bf x} + {\bf b})
\; .
\end{equation}
Because the utility function sums over all symmetry transformations of the feature space, it will also be invariant under all symmetry transformations of the weights ${\hat w} \rightarrow {\hat w} \cdot U(g)$. Thus $\Omega({\hat w}, \ldots) = \Omega({\hat w} U(g), \ldots)$, where $\ldots$ denote any other weights that the model may depend on. This an explicit example of how the degeneracy arises. 

When using symmetry invariant features, we have $\forall g U(g) = 1$, and no degeneracy is forced by symmetry. The utility function will be in the trivial orbit in these cases - the utility function minimum is in general a single point in parameter space. Any degeneracy in the utility function minima will be spurious, and can be broken by any small perturbation to the training dataset. The exact same behavior is found in the physics in the behavior of eigenstates in quantum mechanics. Different eigenstates can be degenerate, and have the same energy level. If the degeneracy is because of a symmetry, then the degeneracy survives perturbations to the system . If the degeneracy is accidental (meaning not due to symmetry), then any perturbation in general will remove the degeneracy.

So far we have discussed a utility function where every possible symmetry transformation of each sample is taken into account. One could imagine sampling this full set of transformed samples, rather than training on its entirety. We can model this random selection as follows
\begin{equation}
\Omega({\bf \theta}) = \sum_i \sum_{g \in G} \eta_{i,g} L(y_i, f(U(g) {\bf x}_i; {\bf \theta}))
\; ,
\end{equation}
where $\eta_{i,g} \in {0,1}$ is a random indicator variable determining whether a certain sample will be taken into account in the training process or not. The probability for each sample to be taken into account should be equal, and therefore there is a homogeneous expectation value $\forall i, g \,\, E(\eta_{i,g}) = \mu$. This causes the expectation value of the utility function to become
\begin{equation}
E \left( \Omega({\bf \theta}) \right) = \mu \sum_i \sum_{g \in G} L(y_i, f(U(g) {\bf x}_i; {\bf \theta}))
\; ,
\end{equation}
thus reverting back to the form of Eq.~\eqref{standard_practice_fitting}.
Much like in disorder physics, while randomness breaks symmetry for any individual instance, the expectation value restores the symmetry.

To close this subsection we note that recent work on deep neural networks identified the two phases of learning~\citep{shwartz1703opening}. The first phase is characterized by a descent into a valley of the utility function. The second is characterized by diffusion within the valley, until a final minimum is settled upon. It is tempting to think that the second diffusive phase may take place in the very same shallow valley we uncover here.

\subsection{General symmetry invariants}

In this subsection we will briefly discuss a general approach towards constructing the symmetry invariant features required to effectively tackle machine learning with symmetry enforced.

Using group theory terminology, every input vector space is a representation of the symmetry group. This representation is a direct sum of irreducible representations (irreps). Each irrep very naturally produces one quadratic invariant of the symmetry group - the norm within that irrep. However, these invariants alone do not generally constitute a full encoding of the information modulo the symmetry transformations. While previous work~\citep{kazhdan2003rotation} using only such invariants has proven this approach useful, we feel the full potential of symmetry enforced upon machine learning models is best served when the encoding is complete, i.e. no information is lost when using symmetry invariant features alone.

\section{Conclusions}\label{Sec:conclusions}

In this article we explored the role symmetry can play in machine learning. We introduced a general way to enforce symmetry onto a machine learning model, namely to preselect features that are invariant under the symmetry transformations.

We demonstrated the dire consequences ignoring symmetry can have on the performance of a constrained family of neural network models, and how enforcing symmetry onto this model can be a benefit, and is not overly complex.

When using symmetry invariant features, we found that the choice of which features to use is significant. Some symmetry invariant features may lose significant amounts of information. In our investigation here, using the feature set $x_i^2$ loses all information about the signs of $x_i$. The two other symmetry invariant feature sets we used retain the sign information either fully in the case of the gradient-like features $x_i x_{i + {\hat e}_1}$ or at least partially in the case of the randomized features $x_i x_{P(i)}$. The significant difference in accuracy between the feature sets that retain the sign information, and the one that does not, indicate the importance of encoding the full input information when constructing symmetry invariant features.

Our examination of the fitting utility function when dealing with symmetry uncovered a danger in the standard approach of fitting the model to transformed copies of the data. The utility function will have symmetry induced degenerate minima. For continuous (Lie) symmetry groups, spurious Goldstone modes (shallow directions in the fit landscape) will arise in this context. Training on symmetry invariant features eliminates these degeneracies. 

In the application of machine learning to image recognition there is a vast amount of symmetry to be found. Translational invariance (which is naturally incorporated in convolutional neural networks), mirror reflections, at least a degree of rotational invariance, scale invariance (zooming in and out), as well as changing the colors themselves while maintaining contrast between different colored pixels are all transformations under which the perception of the image ought to remain unchanged. There is much more work to be done to take full advantage of these symmetries in developing more effective and robust image recognition systems.

\section{Acknowledgements}
The author would like to thank Miles Stoudenmire, Daniel Malinow, David J. Bergman, and Dmitry S. Novikov, for useful feedback on the ideas presented in this manuscript. In particular, discussions with Dmitry S. Novikov inspired exploring the fitting parameter degeneracy
that occurs when symmetry is not enforced upon a fitting model. The author would also like to thank the UCI machine learning repository\footnote{\url{http://archive.ics.uci.edu/ml/index.php}} for making the dataset used in this work available.

\vskip -0.2in

\bibliography{NN_bib}

\vskip 0.2in

\end{document}